\documentclass[10pt,twocolumn,letterpaper]{article}
\pdfoutput=1
\usepackage{cvpr}       
\usepackage{graphicx}
\usepackage{amsmath}
\usepackage{amssymb}
\usepackage{multirow}
\usepackage{booktabs}
\usepackage{url}
\usepackage[accsupp]{axessibility}
\usepackage{algorithm}
\usepackage{algorithmic}

\usepackage[pagebackref,breaklinks,colorlinks]{hyperref}

\newcommand\blfootnote[1]{%
	\begingroup
	\renewcommand\thefootnote{}\footnote{#1}%
	\addtocounter{footnote}{-1}%
	\endgroup
}
\usepackage{tikz}
\definecolor{yellow}{rgb}{1, 1, 0.7}
\definecolor{orange}{rgb}{1, 0.85, 0.7}
\definecolor{tablered}{rgb}{1, 0.7, 0.7}
\definecolor{mypink}{HTML}{E59EDD} % 或用 RGB{229,158,221}
\definecolor{mygray}{HTML}{AEAEAE}

\usepackage[capitalize]{cleveref}
\crefname{section}{Sec.}{Secs.}
\Crefname{section}{Section}{Sections}
\Crefname{table}{Table}{Tables}
\crefname{table}{Tab.}{Tabs.}

\begin{document}
\title{
Hierarchical Flow Matching for 3D Point Cloud Generation
}
\author{
    Linhao Wang $^{1}$, 
    Qichang Zhang $^{2}$, 
    Ye Su $^{1}$,
    Hao Wang $^{1,\dag}$
    \\
    $^{1}$ Shandong Normal University\\
    $^{2}$ University of Macau \\
}

\maketitle
\blfootnote{\dag Corresponding Author}
% \blfootnote{\ddag Project Leader}
% \blfootnote{This work has been submitted to the IEEE for possible publication. Copyright may be transferred without notice, after which this version may no longer be accessible.}

%%%%%%%%% ABSTRACT
\begin{abstract}
Generating high-quality 3D point clouds requires capturing both global shape topology and local geometric details. Existing flow-based methods rely on continuous normalizing flows (CNFs) that demand expensive ODE solving and trace estimation during training, while diffusion models require hundreds of iterative denoising steps. Moreover, most approaches adopt single-level generation directly in point space, disregarding the hierarchical structure natural to 3D shapes. We propose Hierarchical Flow Matching (HFM) that extends flow matching to bilevel structure for unconditional 3D point cloud generation. HFM decomposes the task into two levels via optimal-transport flow matching: a \textit{Latent Flow Matching} models the global shape manifold in a compact latent space, and a \textit{Conditional Point Flow Matching} reconstructs detailed point clouds conditioned on the latent code. Both flows are trained with simple MSE regression losses. The resulting straight OT paths enable efficient sampling with as few as 15 Euler steps per flow, while the structured latent space supports downstream tasks including classification. Extensive experiments on ShapeNet and ModelNet benchmarks demonstrate that HFM achieves competitive or even best performance compared with prior state-of-the-art methods.
\end{abstract}

%%%%%%%%% BODY TEXT
\section{Introduction}
3D point clouds are a fundamental representation in computer vision and graphics, with applications ranging from autonomous driving and robotics to virtual reality. Generative models for point clouds must address two core challenges: the permutation invariance of unordered point sets, and the multi-scale nature of geometric structures that spans global shape topology and local geometric details.

A broad spectrum of generative models has been explored for this task. GAN-based methods~\cite{achlioptas2018learning,gadelha2018multiresolution,gao2022get3d} suffer from training instability and mode collapse, while VAE-based methods~\cite{gadelha2018multiresolution,kim2021setvae} often produce oversmoothed outputs. Diffusion-based methods~\cite{zhou20213d,luo2021diffusion,vahdat2022lion,mo2023dit} achieve high generation quality but require hundreds of iterative denoising steps and lack explicit structured latent representations. Flow-based methods~\cite{kim2020softflow,yang2019pointflow,klokov2020discrete} leverage Normalizing Flows~\cite{papamakarios2021normalizing,dinh2016density,dinh2014nice} or Continuous Normalizing Flows (CNFs)~\cite{chen2018neural} that offer principled density estimation through ODEs but incur substantial training cost, requiring forward and backward ODE solving along with trace estimation~\cite{grathwohl2018ffjord}. Beyond computational efficiency, some of them adopt \textit{single-level generation}, neglecting the hierarchical structure inherent to 3D shapes,
where global topology conditions local geometry.
Some of the methods mentioned above recognized this principle. For instance, PointFlow first adopts a bilevel CNF design but suffers expensive ODE training, while DPM applies a similar decomposition within a diffusion framework yet inherits slow iterative sampling.
In short, both of them realize the bilevel design at the cost of either expensive training (ODE solving) or sampling (many denoising steps). Another line of work~\cite{molodykmfm2025, meng2026pointnsp} addresses coarse-to-fine generation via autoregressive models that discretize shapes into tokens and generate point clouds hierarchically, yet rely on discrete tokenization and autoregressive decoding.

Flow Matching (FM)~\cite{lipman2022flow,liu2022flow,albergo2022building} has recently emerged as a compelling alternative that overcomes both limitations simultaneously, replacing ODE-based training with simulation-free MSE regression.
However, existing FM-based point cloud methods remain single-level: Point Straight Flow (PSF)~\cite{wu2023fast} requires a three-stage pipeline (training, reflow, distillation) for one-step generation,
while Not-So-Optimal Transport Flow~\cite{hui2025not} focuses on OT coupling efficiency, both operate directly in point space without hierarchical factorization. Consequently, the efficiency gains of FM have yet to be combined with a structured, bilevel design for point cloud generation.

In this paper, we propose \textit{Hierarchical Flow Matching (HFM)}, a framework that extends flow matching to bilevel structured generation for unconditional 3D point cloud generation (Fig.~\ref{fig:hfm_framework}). Inspired by PointFlow's insight~\cite{yang2019pointflow}, we replace its computationally expensive CNF backbone with simulation-free flow matching, achieving hierarchy without the ODE burden. Our design is driven by two observations: (i) FM offers dramatically simpler training than CNFs using MSE regression with closed-form targets, no ODE solver or trace estimation; and (ii) hierarchical factorization naturally decomposes point cloud generation into two stages: global latent shape (\textit{what shape?}) and conditional point modeling (\textit{where are the points?}). Concretely, HFM comprises two FM modules trained jointly with an entropy-regularized empirical prior over the latent space:
\begin{itemize}
    \item \textit{Latent Flow Matching} models the global shape prior $p(z)$ via unconditional OT-CFM in a compact latent space.
    \item \textit{Conditional Point Flow Matching} models $p(X|z)$, the point distribution conditioned on the latent code, via conditional OT-CFM.
\end{itemize}
Both flows learn straight OT paths through simple MSE objectives, eliminating ODE solvers, trace estimation, and log-likelihood computation. The structured latent space supports downstream tasks including classification, and the model is trained end-to-end in a single stage without distillation or reflow~\cite{wu2023fast}.

Our contributions are three-fold:
\begin{itemize}
    \item \textbf{Conceptually}, we propose Hierarchical Flow Matching, a bilevel structured flow matching framework for unconditional 3D point cloud generation. This provides a principled decomposition aligned with the natural hierarchy of 3D shapes.
    \item \textbf{Methodologically}, HFM retains a structured latent space, enabling downstream tasks such as classification while eliminating the need for ODE solvers and log-likelihood computation during training. The entire framework is trained end-to-end in a single stage without distillation or reflow.
    \item \textbf{Empirically}, extensive experiments on ShapeNet and ModelNet benchmarks demonstrate that HFM achieves generation quality comparable with or surpassing prior state-of-the-art methods.
\end{itemize}

%-------------------------------------------------------------------------
\section{Related Work}
\label{sec:related}

\subsection{Point Cloud Generation}

Point cloud generation has been extensively studied using GANs~\cite{achlioptas2018learning}, VAEs~\cite{gadelha2018multiresolution,kim2021setvae}, 
diffusion models~\cite{zhou20213d,meng2025lt3sd,luo2021diffusion,vahdat2022lion,mo2023dit,ren2024tiger}, 
and flow-based methods~\cite{yang2019pointflow,kim2020softflow}. 
Diffusion models achieve high generation quality but require hundreds of sampling steps. 
Flow-based methods using CNFs (PointFlow~\cite{yang2019pointflow}, SoftFlow~\cite{kim2020softflow}) offer principled density estimation via neural ODEs at substantial training cost. Specially, PointFlow further extends this to a bilevel architecture with two CNFs. DPM~\cite{luo2021diffusion} similarly adopts a bilevel decomposition within the diffusion framework by conditioning on a shape latent, yet inherits the slow iterative sampling of diffusion models. 
Autoregressive coarse-to-fine models~\cite{molodykmfm2025, meng2026pointnsp} approach point cloud generation via discrete tokenization and hierarchical decoding, which differs fundamentally from our continuous flow matching formulation. 
Flow matching methods (PSF~\cite{wu2023fast}, Not-So-OT~\cite{hui2025not}) and MFM-point~\cite{molodykmfm2025} provide efficient simulation-free training but remain single-level in point space. 
Similar to our work, ArticFlow~\cite{lin2025articflow} also proposes a two-stage flow matching architecture, targeting action-conditioned articulated mechanism generation. Our work differs by focusing on unconditional point cloud generation with an entropy-regularized latent prior, achieving efficient sampling with straight OT paths, and supporting downstream tasks such as classification.

\subsection{Flow-Based Generative Models}

Normalizing Flows~\cite{rezende2015variational,dinh2016density,kingma2018glow} construct complex distributions through a sequence of invertible transformations with tractable Jacobians, yet their discrete architectures limit expressiveness.
Continuous Normalizing Flows (CNFs)~\cite{chen2018neural,grathwohl2018ffjord} generalize this to continuous-time dynamics via neural ODEs, enabling flexible density estimation at the cost of ODE solving and trace estimation during training.
Flow Matching (FM)~\cite{lipman2022flow,liu2022flow} offers a simulation-free alternative by regressing a vector field onto a closed-form conditional target, reducing training to simple MSE regression.
Optimal transport (OT) paths~\cite{pooladian2023multisample,esser2024scaling,tong2023improving} yield straight trajectories for efficient sampling. In this work, we extend the flow matching framework with a bilevel generative structure for point clouds.
\begin{figure*}
    \centering
\includegraphics[width=0.9\textwidth]
    {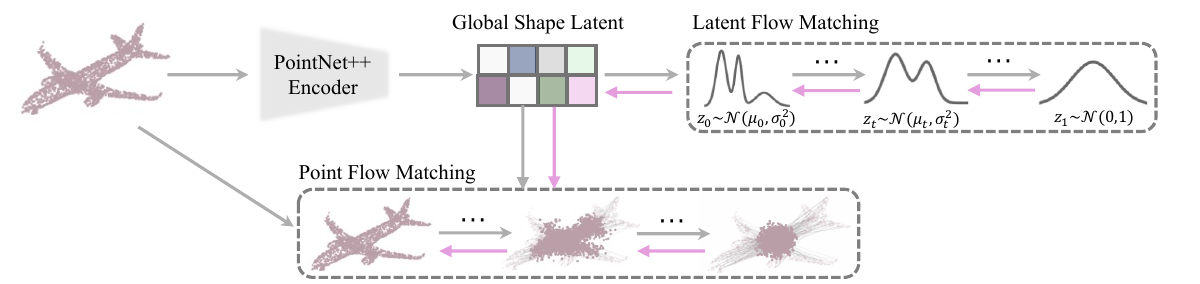}
    \caption{Overview of the proposed Hierarchical Flow Matching (HFM) generation framework consisting of two processes: \textcolor{mygray}{\textit{Training}} and \textcolor{mypink}{\textit{Sampling}}. During training (\textcolor{mygray}{$\rightarrow$}), the encoder $q(z|X)$ maps a point cloud $X$ to a global shape latent $z$, yielding both (i) a Latent Flow Matching objective $\mathcal{L}_{\text{latent}}$ that regresses the velocity field $v_\psi(z_t, t)$ onto the straight OT path from $z$ to noise, and (ii) a Point Flow Matching objective $\mathcal{L}_{\text{point}}$ that regresses $v_\theta(x_t, t, z)$ onto the OT path from $X$ to noise conditioned on $z$. During sampling (\textcolor{mypink}{$\rightarrow$}), Gaussian noise is first transported to a global shape latent by reversing the learned latent flow, and then a second noise sample is transported to a generated point cloud by reversing the point flow conditioned on $z$. Specifically, the thin gray lines represent the movement trajectories of the points.}
    \label{fig:hfm_framework}
\end{figure*}

\section{Method}
\label{sec:method}

\subsection{Overview}
Hierarchical Flow Matching (HFM) decomposes point cloud generation into two levels via optimal-transport flow matching: a \textit{Latent Flow Matching} for the distribution of global shape codes, and a \textit{Conditional Point Flow Matching} for the conditional distribution of point clouds given a shape code. Following the VAE framework~\cite{kingma2013auto}, an encoder~\cite{qi2017pointnet++} maps $X \in \mathbb{R}^{N \times 3}$ to a latent code $z \in \mathbb{R}^d$, and both the prior $p(z)$ and likelihood $p(X|z)$ are modeled as OT flow matching modules (Fig.~\ref{fig:hfm_framework}). 

\subsection{Preliminaries: Conditional Flow Matching}
\label{sec:prelim}

% \subsection{Variational Auto-Encoders}
% VAE~\cite{kingma2013auto} models the data distribution via a latent variable $z$ with a prior $p(z)$, and a decoder $p(X|z)$ that captures the distribution of $X$ given $z$. During training, an encoder $q(z|X)$ approximates the intractable posterior $p(z|X)$. The encoder and decoder are jointly trained by maximizing a variational lower bound on the log-likelihood:
% \begin{equation}
% \log p(X) \geq \mathbb{E}_{q(z|X)}[\log p(X|z)] - D_{\text{KL}}(q(z|X) \parallel p(z)).
% \label{eq:elbo}
% \end{equation}
% In practice, $q(z|X)$ is typically modeled as a diagonal Gaussian $\mathcal{N}(z|\mu_\phi(X), \text{diag}(\sigma_\phi^2(X)))$, and sampling is performed via the reparameterization trick $z = \mu + \sigma \odot \epsilon$ with $\epsilon \sim \mathcal{N}(0, I)$.

Flow Matching (FM)~\cite{lipman2022flow,liu2022flow,albergo2022building} learns a time-dependent vector field $v_\theta: \mathbb{R}^d \times [0,T] \to \mathbb{R}^d$ that transports a noise distribution $p_0$ to a data distribution $p_T$ via an ordinary differential equation. Directly regressing $v_\theta$ onto the marginal vector field is intractable, but Conditional Flow Matching (CFM) circumvents this by conditioning on a data point $x_1 \sim q(x_1)$.

A particularly effective choice is the \textbf{optimal transport (OT) path}, which linearly interpolates between a noise sample $x_0 \sim \mathcal{N}(0,I)$ and a data point $x_1$:
\begin{equation}
x_t = \left(1 - \frac{t}{T}\right) x_0 + \frac{t}{T} x_1,\qquad
u_t(x|x_1) = \frac{x_1 - x_0}{T},
\end{equation}
where $T$ is the final time and $t \in [0,T]$. The CFM objective simplifies to a mean-squared error regression against this closed-form target:
\begin{equation}
\mathcal{L}_{\text{CFM}} = \mathbb{E}_{t, x_0, x_1} \left\| v_\theta(x_t, t) - \frac{x_1 - x_0}{T} \right\|^2.
\label{eq:cfm}
\end{equation}
Training requires no ODE solver, log-likelihood computation, or trace estimation---only a simple MSE loss. The straight OT paths also enable efficient sampling via Euler integration in reverse time:
\begin{equation}
x_{t-\Delta t} = x_t - v_\theta(x_t, t)\,\Delta t.
\end{equation}

\subsection{Latent Flow Matching}

The Latent Flow Matching module models the prior distribution $p(z)$ over shape codes. Since the true prior over shape codes is complex and multi-modal (reflecting diverse categories and structures), we employ unconditional OT-CFM in the latent space to learn a flexible prior without restrictive parametric assumptions.

\noindent \textbf{Latent representation.} The encoder $q(z|X)$ maps a point cloud $X \in \mathbb{R}^{N \times 3}$ to the latent code $z \in \mathbb{R}^d$ via a PointNet++~\cite{qi2017pointnet++} backbone. This compact code captures global topology and category-level structure while delegating fine-grained details to the conditional point flow. Architectural details are provided in the appendix.

\noindent \textbf{Optimal transport flow.} Following the flow matching convention, the OT path interpolates between a shape code $z \sim q(z|X)$ and Gaussian noise $z_0 \sim \mathcal{N}(0, I)$ over time: the interpolant departs from the data at $t=0$ and reaches the noise at $t=T$, following a straight trajectory:
\begin{equation}
z_t = (1 - \tau) z + \tau z_0, \quad \tau = t / T,
\label{eq:latent_path}
\end{equation}
with constant target velocity $u_t = (z_0 - z) / T$, pointing from the data toward the noise. The straight OT paths are especially suitable in latent space: the low-dimensional structure of $z$ means that straight interpolation already provides a near-optimal transport map, allowing the learned flow to efficiently model the manifold of shape codes with few function evaluations.

\noindent \textbf{Objective.} The latent velocity network $v_\psi(z_t, t)$ is trained via MSE regression:
\begin{equation}
\mathcal{L}_{\text{latent}} = \mathbb{E}_{t, z_0, z} \left\| v_\psi(z_t, t) - \frac{z_0 - z}{T} \right\|^2.
\label{eq:latent_loss}
\end{equation}
\begin{figure*}
    \centering
\includegraphics[width=0.92\textwidth]
    {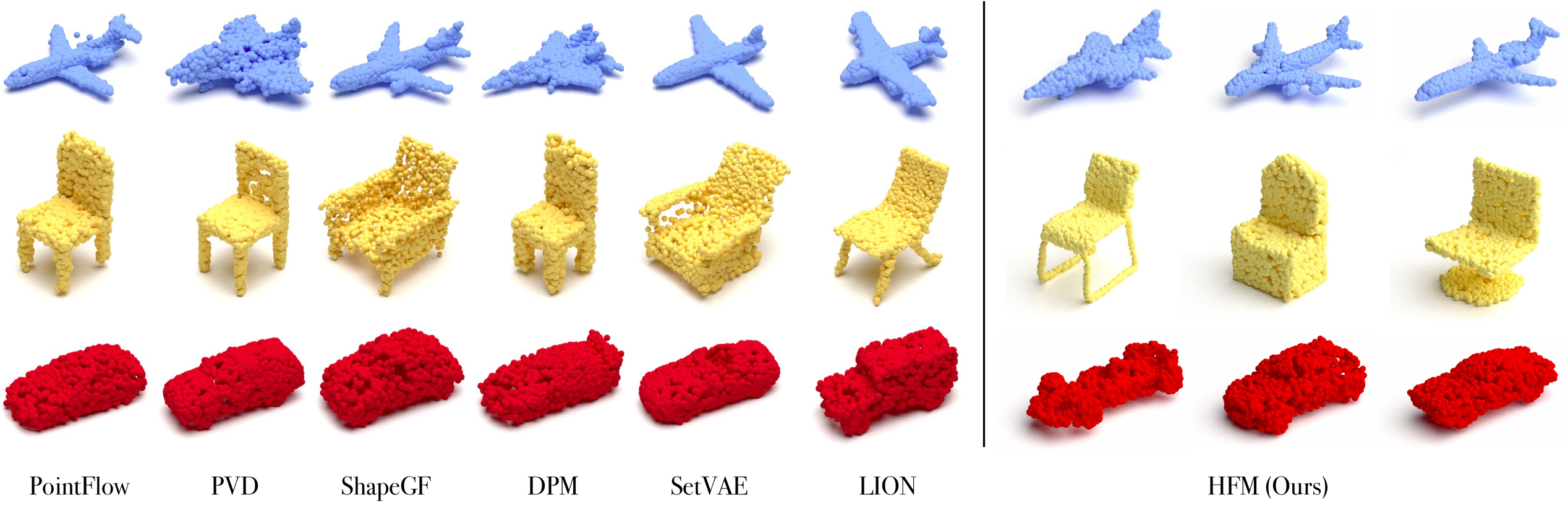}
    \caption{Visualization of our generation results (HFM) compared to baselines.}
    \label{fig:gen_quali}
\end{figure*}

% \noindent \textbf{Design rationale.} A key distinction from standard VAEs is that the latent flow replaces the fixed isotropic Gaussian prior with a learnable distribution. While a standard VAE regularizes the posterior toward $\mathcal{N}(0, I)$ via KL divergence, this fixed prior often mismatches the true distribution of shape codes, particularly when the latent space must capture multi-modal shape variations across categories. The latent flow bridges this gap: it learns $p(z)$ directly from data via simulation-free MSE regression, reducing the regularization burden on the encoder.

\subsection{Conditional Point Flow Matching}

The Conditional Point Flow Matching models $p(X|z)$, the distribution of point clouds conditioned on a shape code. Given the global shape code $z$ encoding coarse topology and category, this module generates the fine-grained geometric details that constitute the complete 3D shape.

\noindent \textbf{Conditional flow formulation.} We employ conditional OT-CFM where the velocity network receives the shape code $z$ as conditioning context. Let $X = \{x^{(i)}\}_{i=1}^N \subset \mathbb{R}^3$ be a point cloud of $N$ points. During training, each point $x^{(i)} \in X$ interpolates toward an independent noise sample $x_0^{(i)} \sim \mathcal{N}(0, I)$:
\begin{equation}
x_t^{(i)} = (1 - \tau) x^{(i)} + \tau x_0^{(i)}, \quad u_t^{(i)} = \frac{x_0^{(i)} - x^{(i)}}{T}, \quad \tau = t / T,
\label{eq:point_path}
\end{equation}
where all points share the same conditioning code $z$, providing global context that coordinates their trajectories into a coherent shape.

\noindent \textbf{Objective.} The conditional velocity network is trained by regressing the OT target. The point velocity network $v_\theta(x_t^{(i)}, t, z)$ operates on each point independently with a shared MLP conditioned on the global context $c = [t, z]$, guaranteeing permutation equivariance by construction. The context $c$ is formed by concatenating the scalar time $t$ with the flattened shape code $z$, and fed into each layer of the MLP via AdaptiveConditionLinear modulation (detailed in the appendix). The objective is analogous to Eq.~\eqref{eq:latent_loss}:
\begin{equation}
\begin{split}
\mathcal{L}_{\text{point}} = \mathbb{E}_{t, x_0^{(i)}, x^{(i)}, z \sim q(z \mid X)}
\bigl\| v_\theta(x_t^{(i)}, t, z) &- \frac{x_0^{(i)} - x^{(i)}}{T} \bigr\|^2,
\end{split}
\label{eq:point_loss}
\end{equation}
where the expectation implicitly samples $i$ uniformly from $\{1, \dots, N\}$. In practice, the loss is averaged over all $N$ points in each cloud.

% \noindent \textbf{Hierarchical decomposition.} A key design insight is that the latent flow and point flow operate at complementary granularities. The former captures global shape identity (``what shape to generate''), while the latter specializes in local geometry (``where to place each point''). The shape code $z$ serves as the information bottleneck between them and retains global information to guide coherent point generation. This hierarchical factorization mirrors the spatial organization of 3D shapes, where global topology constrains but does not fully determine local geometry.

\subsection{Training Objective}

The complete training objective combines three terms:
\begin{equation}
\mathcal{L}_{\text{total}} = \mathcal{L}_{\text{point}} + \mathcal{L}_{\text{latent}} + \mathcal{L}_{\text{entropy}},
\label{eq:total_loss}
\end{equation}
where $\mathcal{L}_{\text{entropy}} = -\lambda_{\text{entropy}} \cdot \mathcal{H}[q(z|X)]$. 
For the diagonal Gaussian encoder $q(z|X) = \mathcal{N}(z|\mu, \operatorname{diag}({\sigma}^2))$, 
the entropy is available in closed form:
\begin{equation}
\mathcal{H}[q(z \mid X)] = \frac{1}{2} \sum_{i=1}^{d} \left( \log \sigma_i^2 + 1 \right) + \text{const},
\end{equation}
which encourages the encoder to maintain a non-degenerate posterior. 
We set $\lambda_{\text{entropy}} = 10^{-4}$ in all generation experiments.

\begin{table*}[t]
\centering
\scalebox{0.91}{
\begin{tabular}{l|cccc|cccc|cccc}
\toprule
\multirow{3}{*}{Method}
& \multicolumn{4}{c}{Chair}
& \multicolumn{4}{c}{Airplane}
& \multicolumn{4}{c}{Car} \\
\cmidrule(lr){2-5} \cmidrule(lr){6-9} \cmidrule(lr){10-13}
& \multicolumn{2}{c}{MMD $\downarrow$}
& \multicolumn{2}{c}{COV(\%) $\uparrow$}
% & JSD $\downarrow$
& \multicolumn{2}{c}{MMD $\downarrow$}
& \multicolumn{2}{c}{COV(\%) $\uparrow$}
% & JSD $\downarrow$
& \multicolumn{2}{c}{MMD $\downarrow$}
& \multicolumn{2}{c}{COV(\%) $\uparrow$} \\
% & JSD $\downarrow$ \\

& CD & EMD & CD & EMD
& CD & EMD & CD & EMD
& CD & EMD & CD & EMD \\
\midrule

r-GAN
& 2.57 & 12.80 & 33.99 & 9.97
& 0.261 & 5.47 & 42.72 & 18.02
& 1.27 & 8.74 & 15.06 & 9.38 \\

l-GAN (CD)
& 2.46 & 8.91 & 41.39 & 25.68
& 0.239 & 4.27 & 43.21 & 21.23
& 1.55 & 6.25 & 38.64 & 18.47 \\

l-GAN (EMD)
& 2.61 & 7.85 & 40.79 & 41.69
& 0.269 & 3.29 & 47.90 & 50.62
& 1.48 & 5.43 & 39.20 & 39.77 \\

DPF-Net
& 2.54 & -- & 44.71 & 48.79
& 0.264 & -- & 46.17 & 48.89
& 1.13 & -- & 45.74 & 49.43 \\

PVD
& 2.62 & -- & 48.84 & 50.60
& 0.224 & -- & \underline{48.88} & \textbf{52.09}
& 1.10 & -- & 41.19 & 50.56 \\

GET3D
& -- & -- & 43.36 & 42.77
& -- & -- & -- & --
& -- & -- & 15.04 & 18.38 \\

SoftFlow
& 2.53 & -- & 41.39 & 47.43
& 0.231 & -- & 46.91 & 47.90
& 1.19 & -- & 42.90 & 44.60 \\

PointFlow
& \underline{2.42} & \underline{7.87} & 46.83 & 46.98
& \underline{0.217} & \underline{3.24} & 46.91 & 48.40
& \underline{0.91} & \underline{5.22} & 44.03 & 46.59 \\

PC-GAN
& 2.75 & 8.20 & 36.50 & 38.98
& 0.287 & 3.57 & 36.46 & 40.94
& 1.12 & 5.83 & 23.56 & 30.29 \\

LION
& -- & -- & \underline{48.94} & \underline{52.11}
& -- & -- & 47.16 & 49.63
& -- & -- & \underline{50.00} & \textbf{56.53} \\

\midrule
HFM (Ours)
& \textbf{2.34} & \textbf{7.82} & \textbf{49.34} & \textbf{52.14}
& \textbf{0.209} & \textbf{3.12} & \textbf{48.91} & \underline{51.14}
& \textbf{0.88} & \textbf{5.13} & \textbf{50.18} & \underline{53.97} \\
\bottomrule
\end{tabular}
}
\caption{Generation results on Airplane, Car and Chair compared with baselines using MMD and COV. MMD-CD and MMD-EMD scores are multiplied by $10^3$ and $10^2$, respectively.}
\label{tab:other_metric_results}
\end{table*}
We consider two training configurations following~\cite{yang2019pointflow,luo2021diffusion}.
For \textit{generation} task, the full objective in Eq.~\eqref{eq:total_loss} is used: both flows are trained jointly end-to-end, and the latent flow learns a flexible prior over the latent space while the entropy penalty encourages the encoder to capture diverse shape variations.
For \textit{auto-encoding} task, we disable the latent flow and entropy terms ($\lambda_{\text{entropy}} = 0$, $\mathcal{L}_{\text{latent}} = 0$), reducing the objective to $\mathcal{L}_{\text{total}} = \mathcal{L}_{\text{point}}$. In this configuration, the model functions as a deterministic auto-encoder where the encoder is supervised solely through gradients from the point flow reconstruction loss, producing latent codes optimized for preserving shape information rather than for generation. The pseudocode for the training procedure is summarized in the appendix.
\subsection{Sampling and Inference}

Given the learned velocity fields, sampling proceeds by reversing the learned flow: the velocity network $v_\psi$ (and $v_\theta$) predicts the direction from data to noise during training, so during sampling we integrate in the opposite direction to transport noise back to data. Concretely, we apply reverse-time Euler integration from $t = T$ to $t = 0$:
\begin{align}
z_{t - |\Delta t|} &= z_t - v_\psi(z_t, t)\, |\Delta t|, \\
x_{t - |\Delta t|} &= x_t - v_\theta(x_t, t, z)\, |\Delta t|,
\label{eq:reverse_euler}
\end{align}
where $|\Delta t| = T / N = -\Delta t > 0$ is the step size. We set $T = 0.5$ in all experiments. The minus sign ($-$) reverses the data-to-noise direction learned during training, so the update moves from noise to data. We also evaluated the higher-order Heun integrator but observed negligible improvement over Euler, confirming that the near-straight OT trajectories introduce minimal truncation error. The pseudocode for sampling is detailed in the appendix.

\begin{figure}
    \centering
\includegraphics[width=1\linewidth]
    {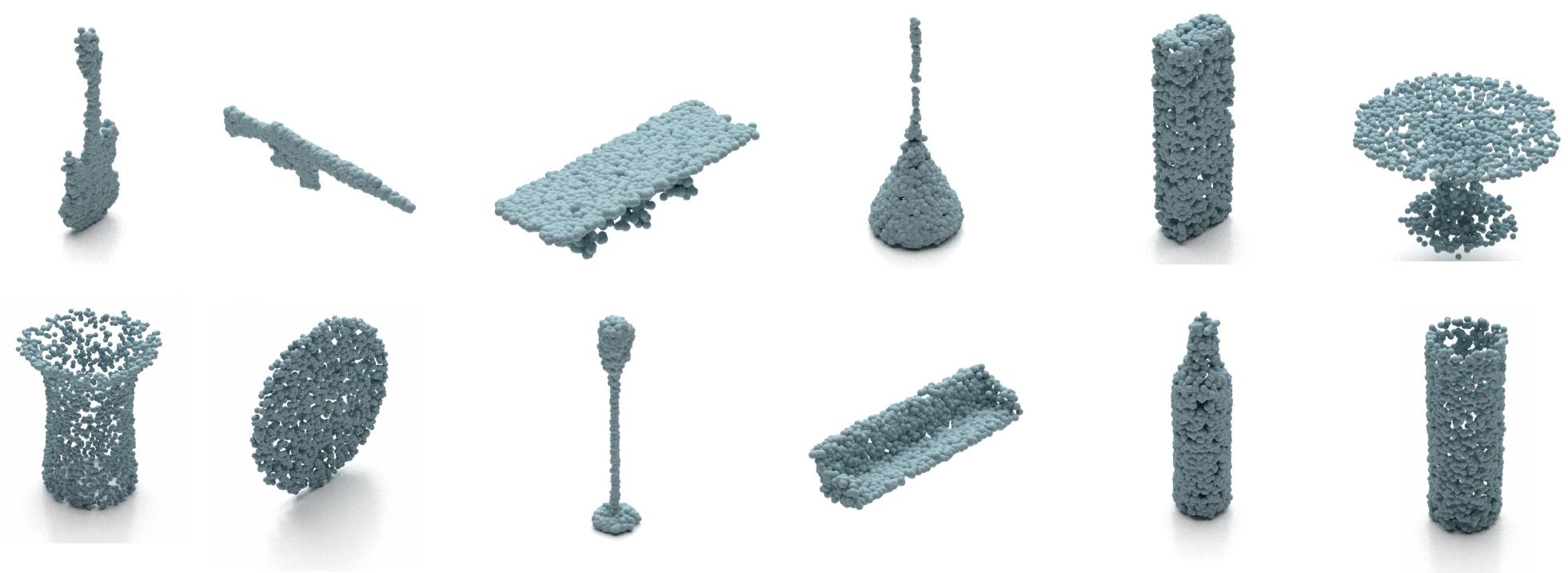}
    \caption{Visualization of multi-class results generated by our method.}
    \label{fig:multiclass_results}
\end{figure}

\section{Experiments}
\label{sec:experiments}
We evaluate HFM on three tasks: unconditional point cloud generation, auto-encoding, and unsupervised representation learning followed by ablation studies. More detailed results and discussion are shown in the appendix.

\subsection{Evaluation Metrics}

Following the evaluation protocols established in prior work~\cite{yang2019pointflow, zhou20213d,vahdat2022lion}, we adopt the following metrics for generation quality:

\noindent \textbf{1-Nearest Neighbor Accuracy (1-NNA).} The leave-one-out accuracy of a 1-NN classifier on the union of generated and reference sets. A score closer to 50\% indicates better quality~\cite{yang2019pointflow}. We report both Chamfer Distance (1-NNA-CD) and Earth Mover's Distance (1-NNA-EMD).

\noindent \textbf{Minimum Matching Distance (MMD).} MMD measures the fidelity of generated point clouds by computing, for each reference shape, the distance to its nearest neighbor in the generated set, then averaging.

\noindent \textbf{Coverage (COV).} COV measures the fraction of reference shapes matched by at least one generated shape as their nearest neighbor. Higher coverage indicates better diversity.

\begin{table}[t]
    \centering
    \scalebox{0.8}{
    \begin{tabular}{lcccccc}
        \toprule
        \multirow{2}{*}{Method} & \multicolumn{2}{c}{Airplane} & \multicolumn{2}{c}{Car} & \multicolumn{2}{c}{Chair} \\
        \cmidrule(lr){2-3} \cmidrule(lr){4-5} \cmidrule(lr){6-7} & CD & EMD & CD & EMD & CD & EMD \\
        \midrule
        r-GAN & 98.40 & 96.79 & 94.46 & 99.01 & 83.69 & 99.70 \\
        l-GAN (CD) & 87.30 & 93.95 & 66.49 & 88.78 & 68.58 & 83.84 \\
        l-GAN (EMD)  & 89.49 & 76.91 & 71.16 & 66.19 & 71.90 & 64.65 \\
        PC-GAN & 94.35 & 92.32 & 92.19 & 90.87 & 76.03 & 78.37 \\
        PointFlow  &  75.68 & 70.74 & 60.65 & 62.36 & 62.84 & 60.57 \\
        SoftFlow  & 76.05 & 65.80 & 62.35 & 54.48 & 59.21 & 60.05 \\
        DPF-Net & 75.18 & 65.55 & 62.35 & 54.48 & 62.00 & 58.53 \\
        SetVAE  & 76.54 & 67.65 & 59.95 & 59.94 & 58.84 & 60.57 \\
        DPM & 76.42 & 86.91 & 68.89 & 79.97 & 60.05 & 74.77 \\
        Shape-GF  & 80.00 & 76.17 & 63.20 & 56.53 & 68.96 & 65.48 \\
        CanonicalVAE  & 80.15 & 76.27 & 63.23 & 61.56 & 62.78 & 61.05 \\
        PVD & 73.82 & 64.81 & 54.55 & 53.83 & 56.26 & 53.32 \\
        GET3D & -- & -- & 75.26 & 72.49 & 75.26 & 72.49 \\
        \midrule
        PointGPT  & 74.85 & 65.61 & 55.91 & 54.24 & 57.24 & 55.01 \\
        PointNSP & 72.24 & 63.29 & \textbf{52.17} & 51.85 & 54.54 & 52.85 \\
        PSF  & 71.11 & \underline{61.09} & 57.19 & 56.07 & 58.92 & 54.45 \\
        LION  & \underline{67.41} & 61.23 & 53.41 & \underline{51.14} & \underline{53.70} & \underline{52.34} \\
        % MFM-point  & \textbf{65.36} & \textbf{57.21} & 57.23 & \textbf{47.87} & \underline{54.92} & 53.25 \\
        SALAD  & 73.90 & 71.10 & 59.20 & 57.20 & 57.80 & 58.40 \\
        NSOT & 68.64 & 61.85 & 59.66 & 53.55 & 55.51 & 57.63 \\
        \midrule
        HFM (Ours) & \textbf{66.61} & \textbf{60.79} & \underline{53.37} & \textbf{51.04} & \textbf{53.62} & \textbf{51.98} \\
        \bottomrule
    \end{tabular}
    }
    \caption{Generation results on Airplane, Car, and Chair compared with baselines using 1-NNA as the metric. Lower is better.}
    \label{tab:1-NN_generation_results}
\end{table}
\noindent \textbf{Classification Accuracy.} For representation learning, we train a linear SVM on the latent codes produced by our encoder and report accuracy on ModelNet datasets.

\subsection{Baselines}

We compare HFM with previous state-of-the-art methods published at top-tier conferences.

For point cloud generation, the baselines are reported in Table~\ref{tab:other_metric_results} and Table~\ref{tab:1-NN_generation_results}, which use complementary metrics: Table~\ref{tab:1-NN_generation_results} reports 1-NNA, while Table~\ref{tab:other_metric_results} reports MMD and COV.
We include GAN-based methods including raw-GAN (r-GAN) and latent-GAN (l-GAN)~\cite{achlioptas2018learning}, PC-GAN~\cite{gadelha2018multiresolution} and GET3D~\cite{gao2022get3d}; VAE-based methods SetVAE~\cite{kim2021setvae}, CanonicalVAE~\cite{cheng2022autoregressive} and ShapeGF~\cite{zhou20213d}; 
Diffusion-based methods including DPM~\cite{luo2021diffusion}, SALAD~\cite{koo2023salad}, PVD~\cite{zhou20213d}, LION~\cite{vahdat2022lion}; 
Flow-based methods such as PointFlow~\cite{yang2019pointflow}, SoftFlow~\cite{kim2020softflow} and DPF-Net~\cite{klokov2020discrete}; 
Flow Matching-based methods PSF~\cite{wu2023fast} and Not-So-OT~\cite{hui2025not}; Autoregressive-based methods PointGPT~\cite{chen2023pointgpt} and PointNSP~\cite{meng2026pointnsp}.

For auto-encoding and representation learning tasks, we compare against latent-GAN (l-GAN)~\cite{achlioptas2018learning}, AtlasNet~\cite{groueix2018papier}, PointFlow~\cite{yang2019pointflow}, and ShapeGF~\cite{zhou20213d} and DPM~\cite{zhou20213d}. Few methods are reported due to the computational costs from training on the whole ShapeNet datasets.

\subsection{Point Cloud Generation}
For unconditional generation, we employ the ShapeNet dataset~\cite{chang2015shapenet} containing 51,127 shapes from 55 categories. Following~\cite{yang2019pointflow}, we train separate models on three representative categories: airplane, chair and car. We sample 2,048 points for training and testing for each shape.
\begin{table}[t]
\centering
\scalebox{0.9}{
\begin{tabular}{lcc}
\toprule
Method & MN-10 (\%) & MN-40 (\%) \\
\midrule
AtlasNet  & 91.9 & 86.6 \\
l-GAN (CD) & \textbf{95.4} & 84.5 \\
l-GAN (EMD)  & \textbf{95.4} & 84.0 \\
PointFlow  & 93.7 & 86.8 \\
ShapeGF & 90.2 & 84.6 \\
DPM & 94.2 & \underline{87.6} \\
\midrule
HFM (Ours) & \underline{94.6} & \textbf{87.9} \\
\bottomrule
\end{tabular}
}
\caption{Comparison of representation learning in linear SVM classification accuracy.}
\label{tab:svm_accuracy}
\end{table}

\noindent \textbf{Qualitative comparisons.} The qualitative results shown in Fig.~\ref{fig:gen_quali} demonstrate the competitive quality generated by HFM compared to baseline methods. The visualization of prior methods is taken from LION~\cite{vahdat2022lion}. Moreover, we also train our model on all 55 different categories from ShapeNet. Fig.~\ref{fig:multiclass_results} shows multi-class point clouds generated by HFM, demonstrating the generalizability of our model.

\begin{table}[t]
\centering
\scalebox{0.9}{
\begin{tabular}{l c}
\toprule
Model & Time (s) \\
\midrule
PointFlow & 0.27 \\
SoftFlow & 0.12 \\
DPM & 22.8 \\
PVD ($N=1000$) & 29.9 \\
PVD-DDIM($N=100$) & 3.15 \\
PSF & \textbf{0.04} \\
\midrule
HFM (Ours) & \underline{0.05} \\
\bottomrule
\end{tabular}
}
\caption{Sampling time comparisons with baselines.}
\label{tab:sampling_time}
\end{table}

\noindent \textbf{Quantitative comparisons.} Table~\ref{tab:other_metric_results} and \ref{tab:1-NN_generation_results} report the full comparisons, and Table~\ref{tab:sampling_time} reports the sampling time comparison when the batch size is one~\cite{wu2023fast}. While PSF is marginally faster than our method, it suffers from a complicated three-stage process. Our method strikes a favorable trade-off between efficiency and effectiveness.
HFM is tested with 15 Euler steps per flow (30 total function evaluations). 

\begin{figure}
    \centering
\includegraphics[width=1\linewidth]
    {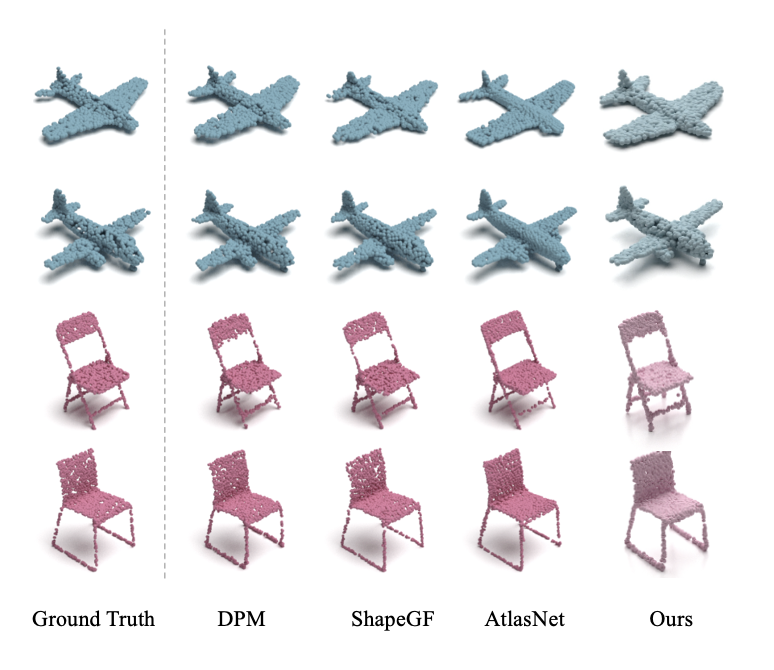}
    \caption{Visualization of our reconstruction results compared with baselines.}
    \label{fig:recon_results}
\end{figure}

\begin{figure*}
    \centering
\includegraphics[width=0.95\textwidth]
    {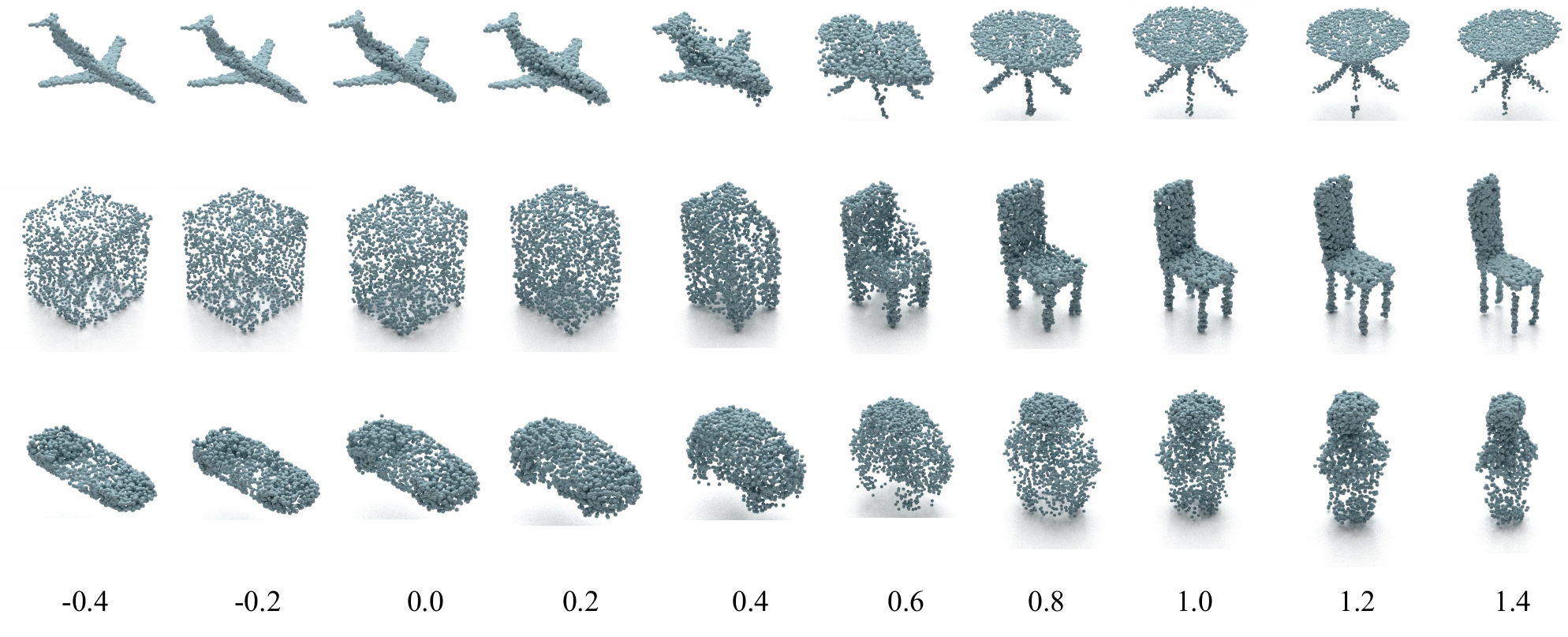}
    \caption{Visualization of interpolation and extrapolation in the latent space.}
    \label{fig:interpolation_results}
\end{figure*}

\subsection{Point Cloud Auto-Encoding}
For auto-encoding, we train an auto-encoder (encoder + Conditional Point Flow Matching) on the full ShapeNet dataset (55 categories) without the latent flow prior. Following~\cite{zhou20213d}, we evaluate reconstruction quality by encoding each test shape and reconstructing via the point flow with 15 reverse Euler steps from random noise. The reconstructed point clouds (2048 points) are compared against the ground truth reference using CD and EMD.

\noindent \textbf{Qualitative comparisons.}  The qualitative comparisons are shown in Fig. \ref{fig:recon_results}, demonstrating competitive quality compared with prior state-of-the-art methods. The qualitative results of previous methods are taken from ~\cite{luo2021diffusion}.

\noindent \textbf{Quantitative comparisons.} As shown in Table~\ref{tab:autoencoding}, HFM achieves the best EMD (3.75) and competitive CD (5.21) compared to baselines. This demonstrates that the Conditional Point Flow Matching, trained with a simple MSE regression objective, can faithfully reconstruct fine-grained geometry from the latent code.
\begin{table}[t]
\centering
\scalebox{0.9}{
\begin{tabular}{l c c}
\toprule
Model & CD & EMD \\
\midrule
AtlasNet  & \textbf{5.13} & 5.97 \\
l-GAN (CD) & 7.12 & 7.95 \\
l-GAN (EMD) & 8.85 & 5.26 \\
PointFlow & 7.54 & 5.18 \\
ShapeGF & 5.73 & 5.05 \\
DPM & 5.25 & \underline{3.78} \\
\midrule
HFM(Ours) & \underline{5.21} & \textbf{3.75} \\
\bottomrule
\end{tabular}
}
\caption{Auto-encoding performance evaluated by CD and EMD. CD and EMD scores are multiplied by $10^4$ and $10^2$, respectively. Lower is better.}
\label{tab:autoencoding}
\end{table}
\begin{figure}[t]
    \centering
    \includegraphics[width=0.8\linewidth]{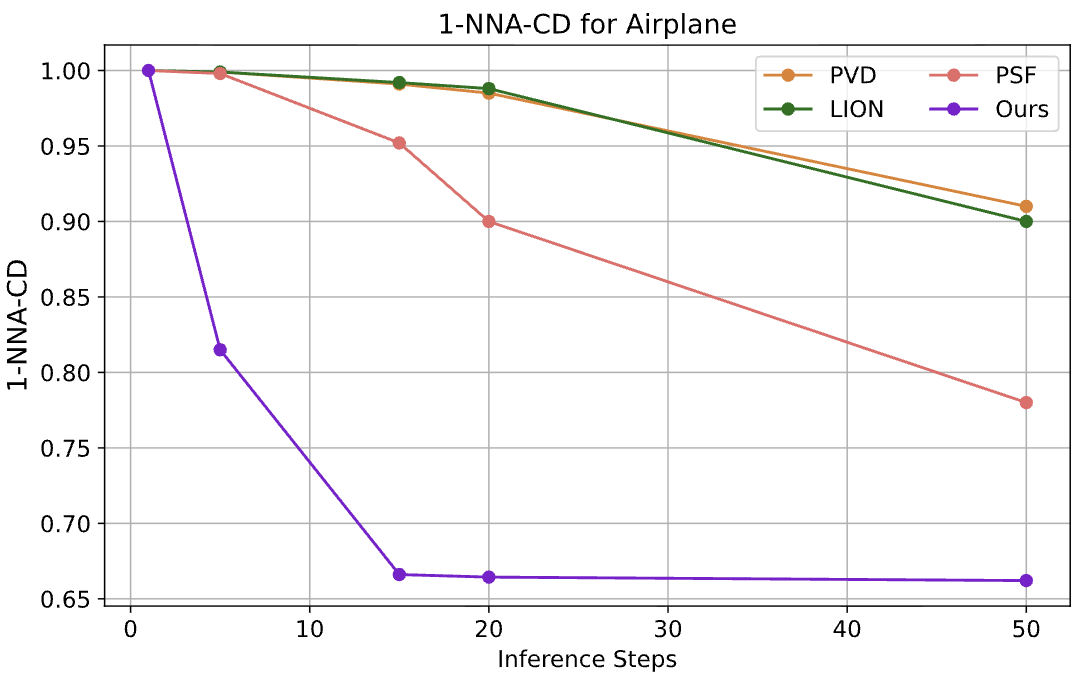} \\
    \includegraphics[width=0.8\linewidth]{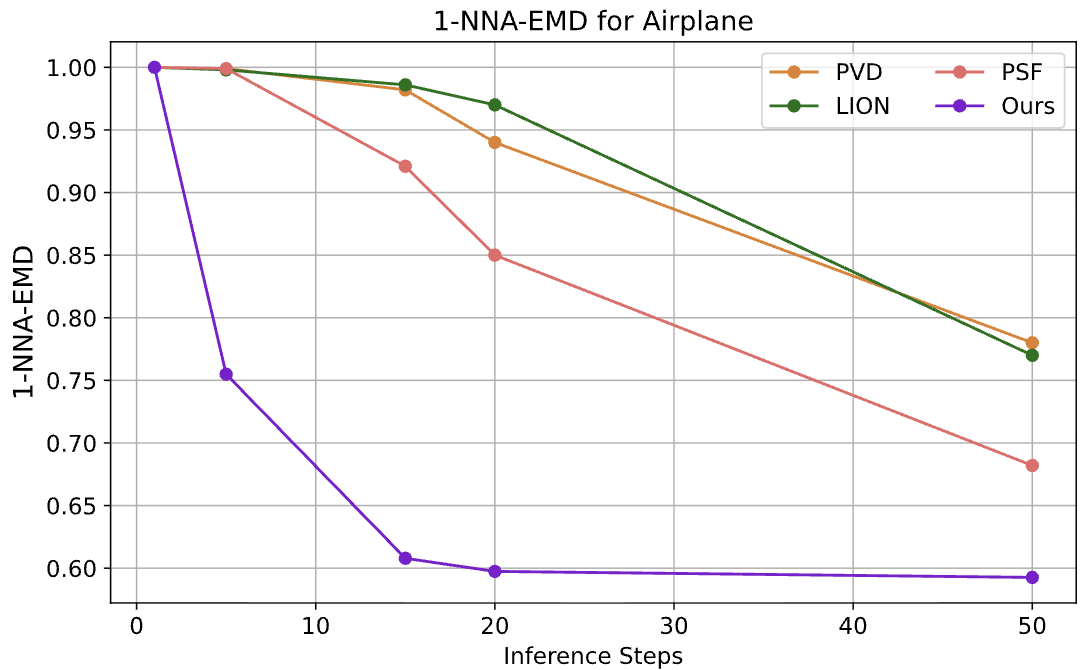} \\
    \caption{Comparison of 1-NNA across different numbers of sampling steps on Airplane category.}
    \label{fig:multistep_results}
\end{figure}
\subsection{Unsupervised Representation Learning}

We evaluate the encoder representations on the ModelNet-10 and ModelNet-40 datasets~\cite{wu20153d}. Following the protocol of~\cite{yang2019pointflow, zhou20213d}, we encode each shape using 10,000 points, train a linear SVM on the training split, and evaluate on the test split. 
As shown in Table~\ref{tab:svm_accuracy}, our method achieves the best MN-40 accuracy (87.9\%) and competitive MN-10 accuracy (94.6\%). Furthermore, we visualize the interpolation and extrapolation in the latent space shown in Fig.~\ref{fig:interpolation_results}.
\subsection{Ablation Studies}

We conduct ablation studies to analyze the latent flow matching, inference steps and entropy coefficient $\lambda_{\text{entropy}}$ on the Airplane category.

\noindent \textbf{Inference Steps.} We compare HFM with representative baselines: LION~\cite{vahdat2022lion}, PVD~\cite{zhou20213d}, PSF~\cite{wu2023fast} without rectified flow by varying the number of inference steps $N \in \{1, 5, 15, 20, 50\}$ on the Airplane category. Fig.~\ref{fig:multistep_results} reports the 1-NNA-CD and 1-NNA-EMD results.
As shown in Fig.~\ref{fig:multistep_results}, PVD, PSF and LION degrade under limited step budgets, whereas HFM maintains robust performance even with as few as 5--15 steps due to the almost straight OT trajectories learned by flow matching.

\noindent \textbf{Latent Flow Matching.} To quantify the contribution of the hierarchical decomposition, we ablate the Latent Flow Matching module on the Airplane category. We compare the full HFM model against a variant where the latent flow is removed (\textit{w/o latent flow}): the encoder still produces a shape code $z$, but during sampling $z \sim \mathcal{N}(0, I)$ is drawn from a standard Gaussian instead of being generated by the learned latent prior. The point flow remains conditioned on $z$ in both cases. \begin{table}[h]
\centering
\scalebox{1}{
\begin{tabular}{lcc}
\toprule
Setting & 1-NNA-CD & 1-NNA-EMD \\
\midrule
w/o latent flow & 78.43 & 74.32 \\
HFM & 66.61 & 60.79 \\
\bottomrule
\end{tabular}
}
\caption{Ablation of the latent flow on Airplane (1-NNA $\downarrow$).}
\label{tab:ablation_latent_flow}
\end{table}
As shown in Table~\ref{tab:ablation_latent_flow}, the learned latent prior provides a better initialization for the conditional point flow compared to sampling $z$ from an uninformative isotropic Gaussian, leading to improved generation quality.

\begin{figure}[h]
    \centering
    \includegraphics[width=0.8\linewidth]{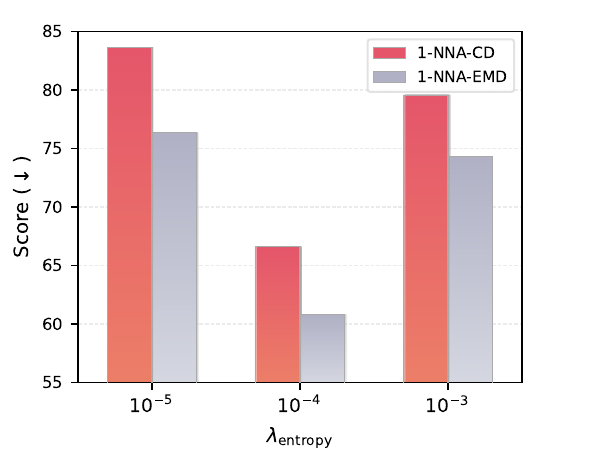}
    \caption{Ablation of $\lambda_{\text{entropy}}$ on Airplane (1-NNA $\downarrow$).}
    \label{fig:ablation_lambda}
\end{figure}

\noindent \textbf{Entropy Coefficient $\lambda_{\text{entropy}}$.} We ablate the entropy coefficient $\lambda_{\text{entropy}}$ on the Airplane category. Fig.~\ref{fig:ablation_lambda} reports 1-NNA-CD and 1-NNA-EMD under varying $\lambda_{\text{entropy}}$. The entropy penalty encourages the encoder to maintain a non-degenerate posterior; too small a value leads to posterior collapse (near-zero variance), while too large a value injects excessive noise that degrades the shape code. We find that $\lambda_{\text{entropy}} = 10^{-4}$ strikes the best balance, achieving the lowest 1-NNA on both metrics.

\section{Conclusion}
\label{sec:conclusion}

We proposed Hierarchical Flow Matching for point cloud generation. HFM is formulated as a Latent Flow Matching and a Conditional Point Flow Matching. With as few as 15 Euler steps per flow, HFM achieves generation quality comparable to or surpassing prior state-of-the-art on ShapeNet and ModelNet benchmarks.

{\small
\bibliographystyle{ieee_fullname}
\bibliography{egbib}
}

\newpage
\begin{figure*}
    \centering
    \includegraphics[width=\textwidth]{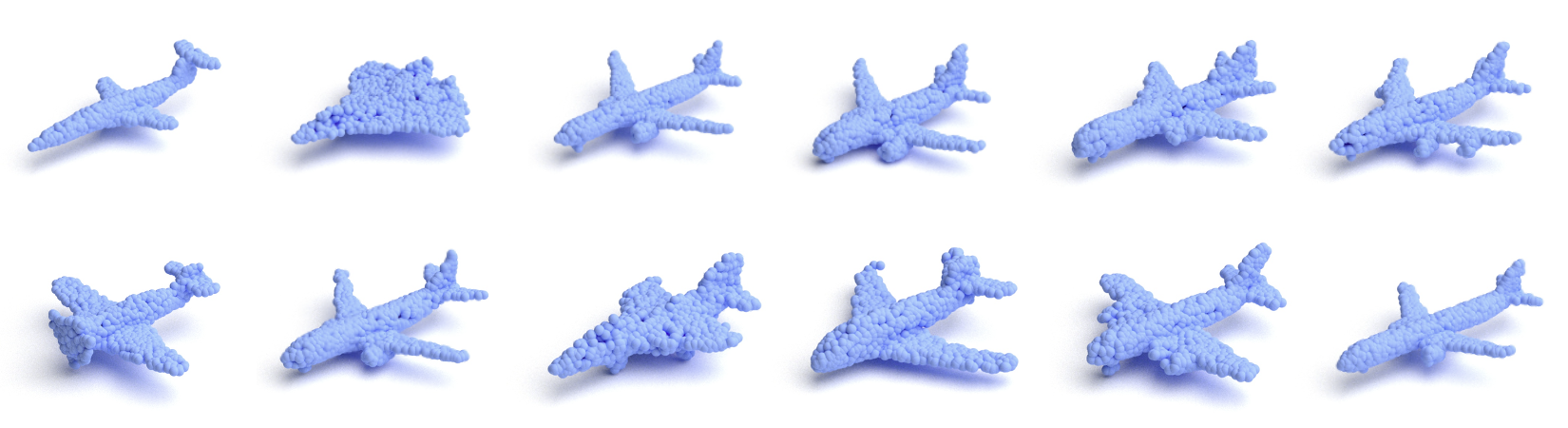}
    \caption{Additional generation results on Airplane.}
    \label{fig:more_airplane}
\end{figure*}
\begin{figure*}
    \centering
    \includegraphics[width=\textwidth]{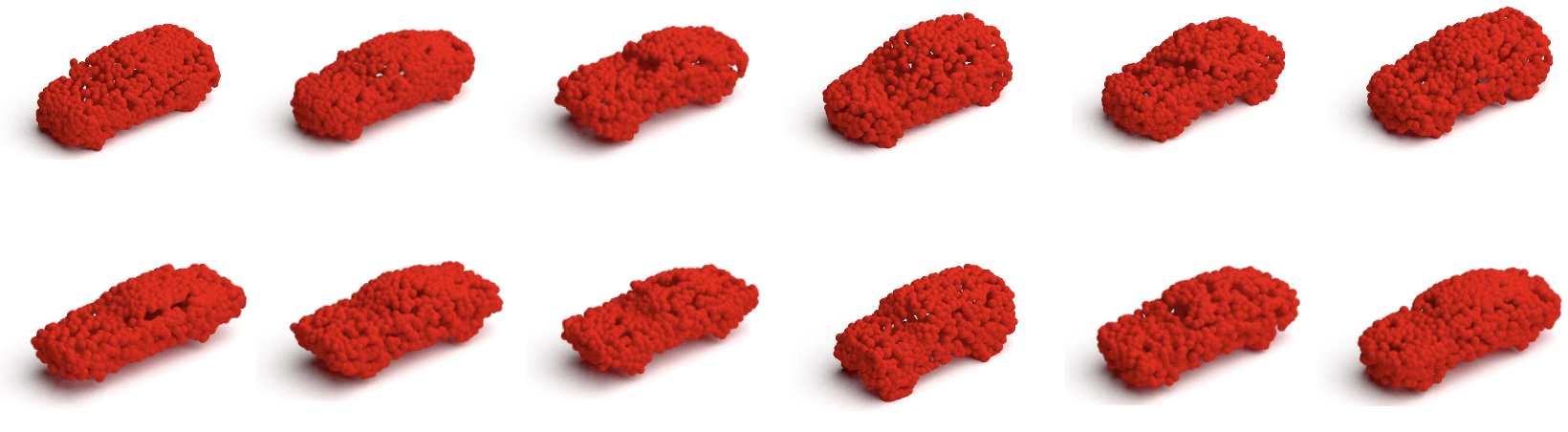}
    \caption{Additional generation results on Car.}
    \label{fig:more_car}
\end{figure*}
\begin{figure*}
    \centering
    \includegraphics[width=\textwidth]{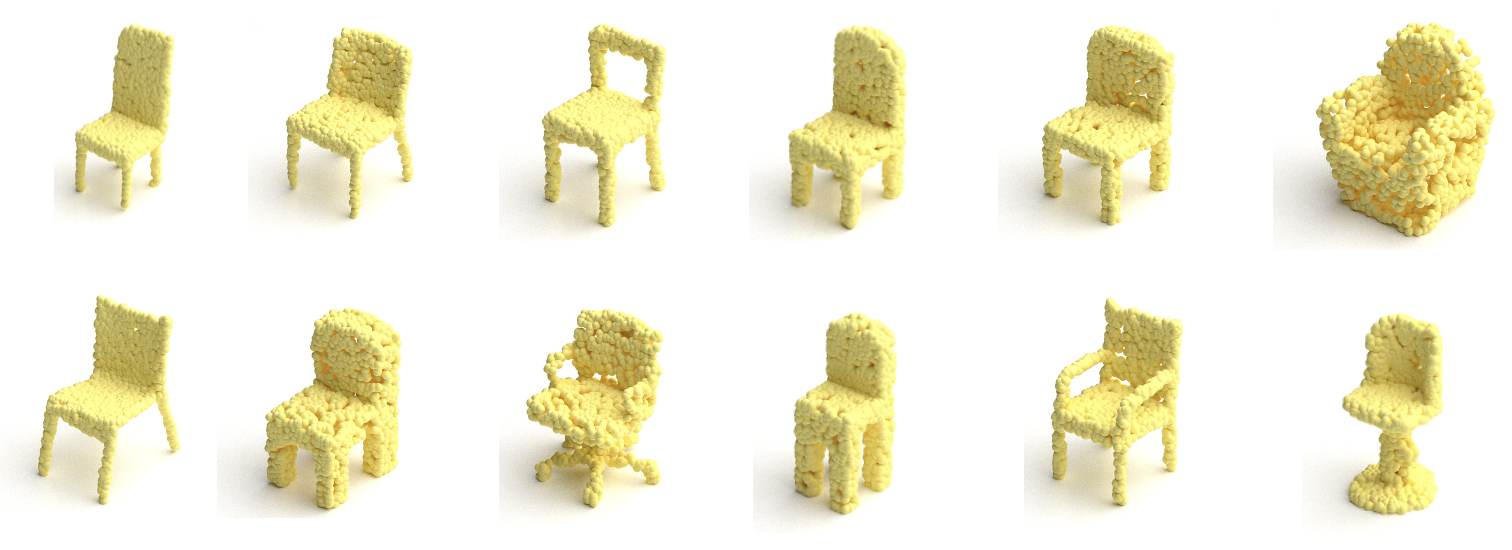}
    \caption{Additional generation results on Chair.}
    \label{fig:more_chair}
\end{figure*}

\section{Implementation Details}
\noindent \textbf{Architecture.} The encoder adopts a PointNet++~\cite{qi2017pointnet++} architecture with three hierarchical Set Abstraction layers: the first downsamples to 512 points via FPS (radius 0.2, 32 neighbors, Conv 3$\rightarrow$64$\rightarrow$128), the second to 128 points (radius 0.4, 64 neighbors, Conv 131$\rightarrow$256$\rightarrow$512), and the third performs global aggregation (Conv 515$\rightarrow$512$\rightarrow$1024), followed by a linear projection (1024$\rightarrow$512) with ReLU and batch normalization to produce a 128-dimensional latent code with diagonal Gaussian parameters $(\mu, \sigma)$, consistent with PointFlow~\cite{yang2019pointflow}.

The point velocity network $v_\theta$ consists of our proposed AdaptiveConditionLinear layers with 256 hidden units and Tanh activations. Given a context vector $c = [t, z_{\text{shape}}]$ (time step concatenated with the shape code), the layer first predicts feature-wise scale $\alpha(c)$ and shift $\beta(c)$ from the context, applies them to modulate the input $\tilde{x} = x \odot (1 + \alpha(c)) + \beta(c)$, then computes the linear transform $W\tilde{x} + b$, and finally adds a context-dependent residual $\gamma(c)$. This design conditions the velocity field on the shape code by modulating the input features, which provides two distinct conditioning pathways: feature recalibration via scale and offset before the weight matrix, and a direct residual connection from the context to the output. The latent velocity network $v_\psi$ uses a similar architecture.

\noindent \textbf{Settings.} We use the Adam optimizer ($\text{lr}=10^{-3}$, $\beta_1=0.9$, $\beta_2=0.999$, no weight decay) with a linear learning rate schedule that decays to zero from the halfway point, batch size of 256, and train on 4 NVIDIA A800-80G GPUs under Linux (Ubuntu) with PyTorch 2.5.1 and CUDA 12.4. The training costs are summarized in Table~\ref{tab:training_compute}. For the per-category setting, we train a separate model on each individual category; the reported GPU hours are averaged across Airplane, Car, and Chair. For the full-dataset setting, we train a single model jointly on all 55 categories, i.e., the reported GPU hours are the total cost of one training run, rather than 55 separate runs. Note that full-dataset training is limited by computational constraints, so per-category experiments on Airplane, Car, and Chair are conducted as the primary benchmarks for comparison with prior works.

\begin{table}[h]
\centering
\caption{Training computational costs.}
\label{tab:training_compute}
\scalebox{0.8}{
\begin{tabular}{lccc}
\toprule
Setting & Task & Epochs & GPU hours \\
\midrule
\multirow{2}{*}{Per-category}
& Generation & $\sim$4.2k & $\sim$9.8 \\
& Auto-encoding & $\sim$4k & $\sim$8.5 \\
\midrule
\multirow{2}{*}{Full-dataset (55 categories)}
& Generation & $\sim$4.1k & $\sim$102 \\
& Auto-encoding & $\sim$4k & $\sim$97 \\
\bottomrule
\end{tabular}
}
\end{table}
\noindent \textbf{Training Objective.} Both $\mathcal{L}_{\text{point}}$ and $\mathcal{L}_{\text{latent}}$ receive gradients with respect to the reparameterized latent code $z$ ~\cite{kingma2013auto}, and thus backpropagate through $z$ to the encoder parameters $(\mu, \log \sigma)$ via the reparameterization trick~\cite{kingma2013auto}. This couples the encoder with both flow modules: the encoder is encouraged to produce latents that are not only informative for point reconstruction (via $\mathcal{L}_{\text{point}}$) but also compatible with the latent flow prior (via $\mathcal{L}_{\text{latent}}$).

\noindent \textbf{Pseudocode.} We provide pseudocode for all four configurations of HFM. Algorithm~\ref{alg:training_gen} summarizes the full training procedure for the generation task, where the encoder, latent flow, and point flow are jointly optimized with the entropy-regularized objective. Algorithm~\ref{alg:training_ae} describes the auto-encoding training, where the latent flow and entropy terms are disabled, reducing the objective to $\mathcal{L}_{\text{point}}$ only. Algorithm~\ref{alg:sampling_gen} outlines the two-stage sampling procedure for generation. Algorithm~\ref{alg:recon_ae} shows the reconstruction procedure for the auto-encoding setting, where $z$ is obtained by encoding a given shape and only the point flow reverse is performed.

\section{More Generation Results}
\label{sec:more_results}
We provide additional qualitative generation results for Airplane, Car, and Chair categories in Fig.~\ref{fig:more_airplane}, Fig.~\ref{fig:more_car}, and Fig.~\ref{fig:more_chair}, respectively.

\section{Limitations and Future Work}
\label{sec:limitations}
\noindent \textbf{Point-wise independent OT coupling.} In the current Conditional Point Flow Matching, each point independently follows an OT path from noise to data, without explicit inter-point coupling beyond the shared shape code $z$. While this design is simple and effective, recent studies on minibatch OT coupling~\cite{pooladian2023multisample, tong2023improving} suggest that coupling noise across points within a shape could further improve local structure coherence and coverage. We leave exploration of such coupling strategies to future work.

\noindent \textbf{MLP-based velocity network.} Our current velocity network uses MLP with AdaptiveConditionLinear layers, which processes each point independently. While permutation-equivariant and effective, MLPs are less expressive than Transformer-based architectures~\cite{vaswani2017attention} in capturing long-range dependencies across points. Recent work~\cite{mo2023dit, ren2024tiger} has demonstrated the benefits of Transformer backbones for point cloud generation. Exploring Transformer-based velocity networks within the flow matching framework is a promising direction for further improving generation quality.

\newpage
\begin{algorithm}[tb]
\caption{HFM Training (Generation)}
\label{alg:training_gen}
\begin{algorithmic}[1]
\STATE \textbf{Input:} Dataset $\mathcal{D}$, encoder $\text{Enc}$, latent flow $v_\psi$, point flow $v_\theta$
\FOR{iteration $=1$ \TO max\_iter}
    \STATE Sample minibatch $X \sim \mathcal{D}$
    \STATE $\mu, \log\sigma \gets \text{Enc}(X)$
    \STATE $z \sim \mathcal{N}(\mu, \text{diag}(\sigma^2))$ \hfill \textit{\# reparameterize}
    \STATE $\mathbb{H} \gets \tfrac{1}{2}\sum(\log\sigma^2) + \text{const}$ \hfill \textit{\# entropy}
    \STATE Sample $t \sim \text{Uniform}(0, T)$, $z_0 \sim \mathcal{N}(0, I)$
    \STATE $z_t \gets (1 - t/T)\,z + (t/T)\,z_0$
    \STATE $\mathcal{L}_{\text{latent}} \gets \| v_\psi(z_t, t) - (z_0 - z)/T \|^2$
    \STATE Sample $t \sim \text{Uniform}(0, T)$, $x_0 \sim \mathcal{N}(0, I)$
    \STATE $x_t \gets (1 - t/T)\,X + (t/T)\,x_0$
    \STATE $\mathcal{L}_{\text{point}} \gets \| v_\theta(x_t, t, z) - (x_0 - X)/T \|^2$
    \STATE $\mathcal{L} \gets \mathcal{L}_{\text{point}} + \mathcal{L}_{\text{latent}} - \lambda_{\text{entropy}}\,\mathbb{H}$
    \STATE Backpropagate $\mathcal{L}$
\ENDFOR
\end{algorithmic}
\end{algorithm}

\begin{algorithm}[tb]
\caption{HFM Training (Auto-Encoding)}
\label{alg:training_ae}
\begin{algorithmic}[1]
\STATE \textbf{Input:} Dataset $\mathcal{D}$, encoder $\text{Enc}$, point flow $v_\theta$
\FOR{iteration $=1$ \TO max\_iter}
    \STATE Sample minibatch $X \sim \mathcal{D}$
    \STATE $\mu, \log\sigma=0 \gets \text{Enc}(X)$
    \STATE $z \gets \mu$ \hfill \textit{\# deterministic}
    \STATE Sample $t \sim \text{Uniform}(0, T)$, $x_0 \sim \mathcal{N}(0, I)$
    \STATE $x_t \gets (1 - t/T)\,X + (t/T)\,x_0$
    \STATE $\mathcal{L}_{\text{point}} \gets \| v_\theta(x_t, t, z) - (x_0 - X)/T \|^2$
    \STATE Backpropagate $\mathcal{L}_{\text{point}}$
\ENDFOR
\end{algorithmic}
\end{algorithm}

\begin{algorithm}[tb]
\caption{HFM Reconstruction (Auto-Encoding)}
\label{alg:recon_ae}
\begin{algorithmic}[1]
\STATE \textbf{Input:} Data point $X$, encoder $\text{Enc}$, point flow $v_\theta$, steps $N_p$
\STATE $\mu, \log\sigma=0 \gets \text{Enc}(X)$ \hfill \textit{\# encode}
\STATE $z \gets \mu$ \hfill \textit{\# deterministic}
\STATE Sample $x \sim \mathcal{N}(0, I)$ \hfill \textit{\# point noise}
\FOR{$j = 1$ \TO $N_p$}
    \STATE $t \gets T - j \cdot T / N_p$
    \STATE $x \gets x - v_\theta(x, t, z) \cdot T / N_p$ \hfill \textit{\# reverse point flow}
\ENDFOR
\STATE \textbf{return} $x$
\end{algorithmic}
\end{algorithm}

\begin{algorithm}[tb]
\caption{HFM Sampling (Generation)}
\label{alg:sampling_gen}
\begin{algorithmic}[1]
\STATE \textbf{Input:} Latent flow $v_\psi$, point flow $v_\theta$, steps $N_l$, $N_p$
\STATE Sample $z \sim \mathcal{N}(0, I)$ \hfill \textit{\# latent noise}
\FOR{$i = 1$ \TO $N_l$}
    \STATE $t \gets T - i \cdot T / N_l$
    \STATE $z \gets z - v_\psi(z, t) \cdot T / N_l$ \hfill \textit{\# reverse latent flow}
\ENDFOR
\STATE Sample $x \sim \mathcal{N}(0, I)$ \hfill \textit{\# point noise}
\FOR{$j = 1$ \TO $N_p$}
    \STATE $t \gets T - j \cdot T / N_p$
    \STATE $x \gets x - v_\theta(x, t, z) \cdot T / N_p$ \hfill \textit{\# reverse point flow}
\ENDFOR
\STATE \textbf{return} $x$
\end{algorithmic}
\end{algorithm}

\end{document}